# Developing Knowledge-enhanced Chronic Disease Risk Prediction Models from Regional EHR Repositories

Jing Mei, PhD[1], Eryu Xia, PhD[1], Xiang Li, PhD[1], Guotong Xie, PhD[1]
[1]IBM Research, Beijing, China

**Abstract**

*Precision medicine requires the precision disease risk prediction models. In literature, there have been a lot well-established (inter-)national risk models, but when applying them into the local population, the prediction performance becomes unsatisfactory. To address the localization issue, this paper exploits the way to develop knowledge-enhanced localized risk models. On the one hand, we tune models by learning from regional Electronic Health Record (EHR) repositories, and on the other hand, we propose knowledge injection into the EHR data learning process. For experiments, we leverage the Pooled Cohort Equations (PCE, as recommended in ACC/AHA guidelines to estimate the risk of ASCVD) to develop a localized ASCVD risk prediction model in diabetes. The experimental results show that, if directly using the PCE algorithm on our cohort, the AUC is only 0.653, while our knowledge-enhanced localized risk model can achieve higher prediction performance with AUC of 0.723 (improved by 10.7%).*

**Introduction**

Generally, development of a chronic disease risk prediction model needs years of studies, and a famous example is the Framingham Heart Study, which began in 1948 and is now on its third generation of participants[1]. Such high-quality evidence is undoubtedly expected to be guidance for applications in clinical decision support. However, when applying those (inter-)national risk models to a localized population, e.g. at a tier II city in China, the prediction performance is not satisfactory. Well, is it possible to develop a localized chronic disease risk prediction model for each regional population (e.g. as of June 2016, China has 662 cities)?

To address this problem, it requires two important keys. First is the data, and for traditional clinical research, the data consists of long-term follow-up questionnaires. We cannot expect to call for participants in each regional population, but thanks to development of regional health information systems in China, Electronic Health Record (EHR) repositories have been growing up with a great amount of longitudinal patient visits[2]. Thus, the above question is now decomposed into two, and the first sub-question is: (1) Does an EHR repository has the equal data quality for developing a chronic disease risk prediction model? We realize this sub-question has been positively answered in related work[3,4,5], but frankly, when exploring the EHR data of a developing country, e.g. China, we have more data quality issues, and we will systematically present the problems and possible solutions in this paper.

Then is the technique, and in literature, either traditional clinical research algorithms (such as logistic regression and cox regression), or advanced machine learning algorithms (such as neural networks and reinforcement learning), have more or less proven their prediction power[3,4,5]. However, there is few work on the localization of (inter-)national chronic disease risk prediction models, by deep learning from regional EHR repositories. Those well-established models are often treated as the competitors (e.g. for the baseline), but not the cooperators. In this paper, we are not going to reinvent the wheels, and instead, we are trying to answer the second sub-question: (2) How to develop a localized chronic disease risk prediction model by injecting domain knowledge into the EHR data learning process?

Actually, the idea of knowledge combination with learning is not novel, and in a high level perspective, the field of artificial intelligence itself is a combination of knowledge representation/reasoning and machine learning/inference. Interest in this combination could trace back to the 1990s, and one of the representatives was the Knowledge-based Artificial Neural Networks (KBANN)[6], mapping propositional logic into neural networks. Later, Markov Logic Networks (MLN)[7] were proposed, combining first-order logic and probabilistic graphical models. For applications, the work of Fold.all demonstrated its improved performance in topic discovery from movie reviews and PubMed abstracts[8]. Another recent work of teacher-student networks presented its success on a knowledge-harnessed convolutional neural network for sentiment analysis, and a knowledge-harnessed recurrent neural network for named entity recognition[9]. More interestingly, even physics knowledge, such as Kinematic equations and free fall, haven been used to supervising neural networks[10]. However, in healthcare domain, the knowledge learning symbiosis still needs to exploit its way. Particularly, we observe different knowledge types in clinical problems, e.g. logical rules (like if-then) for treatment recommendations, arithmetic formulas (like logistic regression and cox regression) for risk prediction, knowledge graph (like SNOMED ontology) for feature correlation. Besides, there are different knowledge injection methods, and in terms of learning process, we roughly category them into three types (which formal definitions will be presented in the section of model learning): (1) knowledge injection to input features, (2) knowledge injection to objective functions, (3) knowledge injection to output labels. For the first injection type, a straightforward implementation is to add knowledge-based features, e.g. those well-known risk factors and/or the calculated risk scores in a disease risk model. For the second injection type, most related work used logic constraints as regularization terms in the learning objective function, when training neural networks (e.g. the teacher-student networks[9]) or probabilistic graphical models (e.g. the Fold.all framework[8]). The latest work directly specifies the knowledge constraints as objective functions, without direct examples of input-output pairs[10]. For the third injection type, it could be a decision fusion of knowledge-generated labels and data-generated labels[11], and such decision fusion could be majority voting, weighted average, meta-classification, etc. In this paper, we focus on the risk models as our knowledge type, and implement all of the three knowledge injection methods, where our knowledge-enhanced neural networks (KENN)[12] is one of the implementations for knowledge injection to objective functions.

For experiments, we start from a raw regional EHR repository, and by relational association with patient IDs and encounter IDs, we construct a cohort for predicting Atherosclerotic Cardiovascular Disease (ASCVD) in diabetes. This is a clinically meaningful problem, special for China, who has 114 million adults suffering from diabetes, by 2013[13]. However, the well-known ASCVD risk models, like Pooled Cohort Equations (PCE)[14], are almost developed in American and/or European. Imaginably, when applying this PCE model in our cohort, we only get a moderate prediction performance, i.e. the area under the receiver operating characteristic curve (AUC) is 0.653. Next, we explore learning algorithms to combine this PCE model with our EHR data analysis, and our experimental results show higher prediction performance, and the AUC reaches to 0.723, improved by 10.7%.

**Methods**

Figure 1 shows our pipeline to develop a knowledge-enhanced chronic disease risk prediction model from regional EHR repositories. Except the knowledge part, as scoped by dashed lines, the pipeline has been introduced and validated in previous work[15]. In the pipeline, first is a clear problem definition, including the outcome and population of interest. Then, we will extract data from an EHR repository to build feature vectors, and due to the EHR data quality issues (to be discussed below), we possibly need to do data cleansing and missing data imputation. After that, feature selection would be applied to find the potential risk factors. Meanwhile, from well-established risk models, we can construct those known risk factors, and calculate the risk scores. Next, we will train prediction models using different algorithms, and in particular, we will introduce our knowledge-enhanced neural networks[12]. Finally, we will evaluate their prediction performance in term of AUC by train/test splitting.

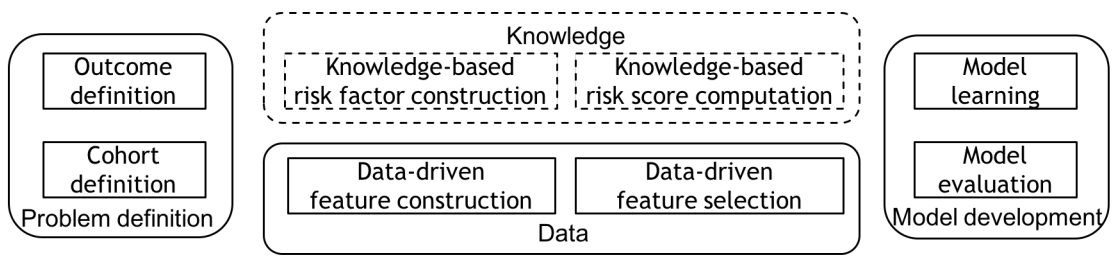

**Figure 1.** Pipeline to develop knowledge-enhanced chronic disease risk models from regional EHR repositories.

Considering that we are facing a raw regional EHR repository, which consists of dozens of relational tables, this section will firstly introduce our EHR data schema, followed by pipeline components.

**EHR data schema.** Figure 2 shows a simplified data schema of our EHR repository. Its core is an Encounter table, with primary key of encounter id. Given an encounter, the information of who (patient id), where (organization id), when (commit time), and how (icd code, diagnosis and cost) are all recorded in the Encounter table. Also, the encounter type includes outpatient, inpatient, and follow-up visits etc. By foreign key of encounter id, other tables of Medication, LabTest, PhyExam, and FollowUp, etc. are associated. Demographic information is documented in the Patient table, with patient id mapping to the Encounter table. Since that's a regional EHR repository, hospital names are documented in the Organization table, with organization id mapping to the Encounter table.

It is remarkable that in China, the Steering Committee of Health Informatics led by Ministry of Health published the "Guideline for Building the Regional Health Information Platform based on Electronic Health Records" in 2009[16]. This guideline describes a technical architecture, which includes a regional health information platform to exchange data among disparate healthcare information systems and a central EHR data repository that stores a longitudinal electronic record of patient health information generated by encounters in any care delivery setting. Since then, regional health information platforms have been developed and deployed almost in every city in China[17], and the EHR repository described in this paper was one of them. In this respect, Figure 2 illustrates a common EHR data schema with Chinese characteristics while following international standards of HL7 CDA and IHE XDS.

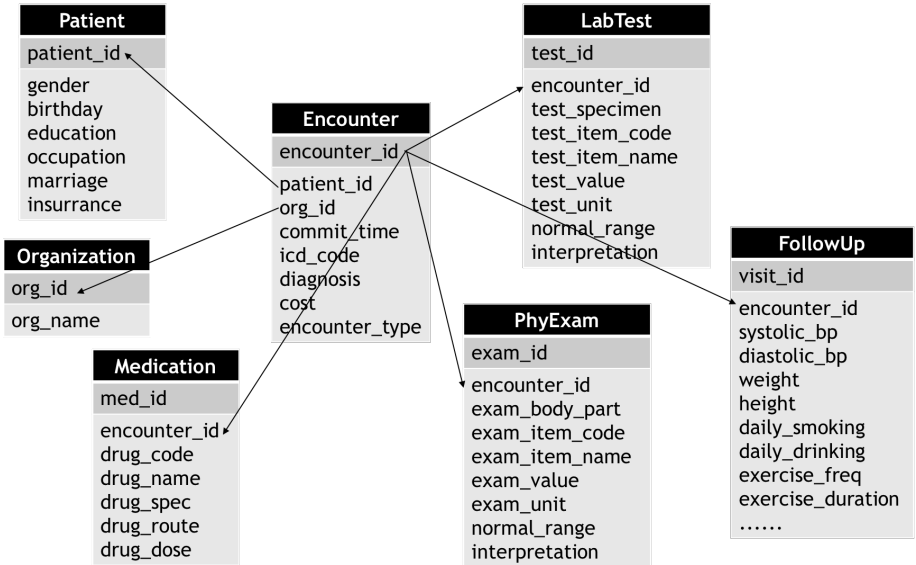

**Figure 2.** A simplified data schema of a regional EHR repository.

**Outcome definition.** ASCVD – defined as acute coronary syndromes (ACSs), a history of myocardial infarction (MI), stable or unstable angina, coronary or other arterial revascularization, stroke, transient ischemic attack, or peripheral arterial disease presumed to be of atherosclerotic origin – is the leading cause of morbidity and mortality for individuals with diabetes and is the largest contributor to the direct and indirect costs of diabetes[18]. A precision ASCVD risk prediction model in diabetes is critical for secondary prevention to improve outcome and reduce cost.

The technical issue is how to identify which patients with/without ASCVD from the EHR data perspective. Generally, for registry data, it is often to use questionnaires to collect data, and e.g., via a checkbox, the ASCVD outcome is easily identified by clicking the checkbox or not. Differently, in EHR, patient data is collected in terms of encounters, and the ASCVD outcome identification needs to check the sequence of encounters for a given patient. As shown in Figure 2, the Encounter table has columns of icd_code and diagnosis. A design purpose of the icd_code column is to regularize the usage of international classification of diseases, i.e. ICD-10 code system[19], but unfortunately, in real clinical settings, it exists null values and a lot invalidate values. By statistics, in the Encounter table, 6.7% is null, and among those non null values, only 54.7% can match the standard ICD-10 codes. The other column of diagnosis is designed for free text, and in our EHR repository, ASCVD is often not documented as a disease name, but a collection of disease names and even the subclasses. Worse, there are quite a few Chinese synonyms for every disease name in ASCVD, and for instance, stroke has at least 3 Chinese names, with dozens of subclasses. After consultation with local domain experts (esp. depending on their medical recording styles), we finally use fuzzy string match in the diagnosis column of the Encounter table, and find 334 distinct text values as identified for ASCVD. In the future work, we consider using advanced natural language processing technologies for a better medical concept learning[20].

**Cohort definition.** The population of interest is a cohort of type 2 diabetes in adults (i.e. age >= 18) without ASCVD at the baseline. As shown in Figure 3, the cohort was constructed using de-identified patient records from a regional EHR repository. All medical records of diabetic patients treated in the index period between August 1st, 2012 and March 31st, 2016 (i.e., 44 months) were extracted. The index event was defined as a type 2 diabetes mellitus (T2DM) related diagnosis, prescription for T2DM medication, or both occurring at any time on or after August 1st, 2012 until the end of study follow-up. Included patients also had at least one medical visit within 360 days before the index period, and at least one T2DM outpatient visit from the index event to the end date – these two inclusive criteria were defined to ensure patients observable in the study period (special for a retrospective cohort extracted from EHR). Patients were excluded if they were diagnosed with gestational diabetes or type I diabetes, were under 18 years old, or if their gender was unknown. After applying these criteria, 89,558 T2DM patients were eligible for inclusion.

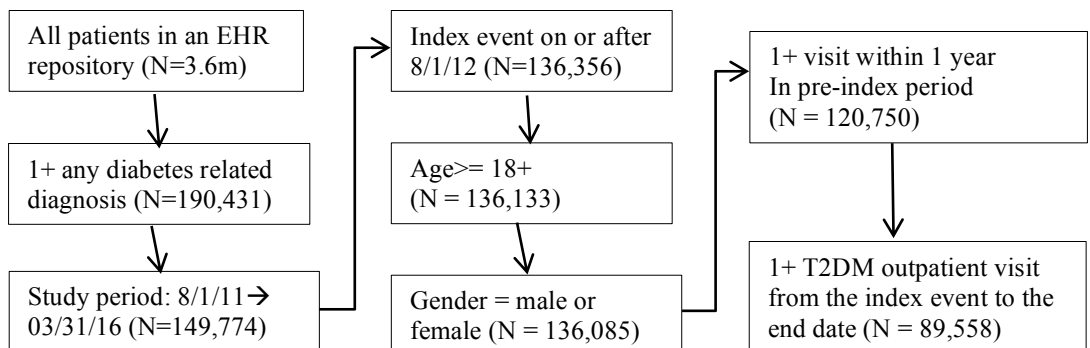

**Figure 3.** Inclusive criteria for the cohort definition from an EHR repository.

Based on the 89,558 included patients, we furthermore divided them into three groups: (1) ASCVD diagnosis observed before T2DM diagnosis; (2) ASCVD diagnosis observed after T2DM diagnosis; (3) ASCVD diagnosis not observed. Figure 4 shows the counting number for each group, and finally, the population of interest stands for individuals in groups (2) and (3). The total number of our cohort is now 13,419 + 41,063 = 54,482, with an observation time window of 1 year before the index event (i.e. T2DM diagnosis firstly observed) to predict the risk of ASCVD.

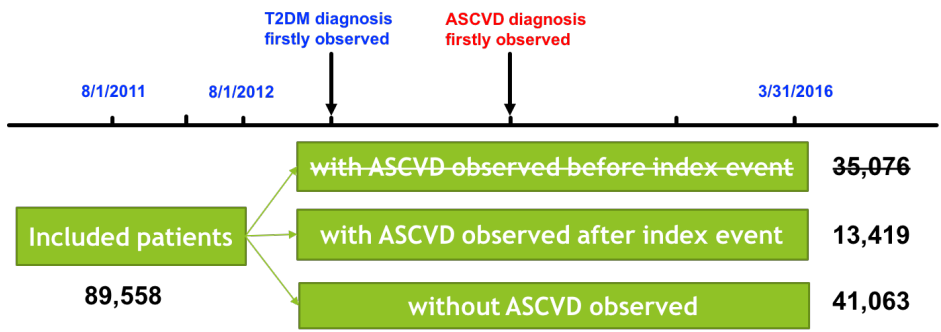

**Figure 4.** Time window and groups of included patients.

**Knowledge-based risk factor construction.** The Pooled Cohort Equations (PCE) has been recommended in ACC/AHA guidelines to estimate the 10-year risk of ASCVD among patients who have never had one of those ASCVD events in the past[21]. Undoubtedly, this knowledge could be leveraged as the seed for developing our localized ASCVD risk prediction model in diabetes. As mentioned in Introduction, the knowledge injection could have different implementation methods, and the straightforward one is to add knowledge-based risk factors and/or the calculated risk scores. Thus, we need to construct the knowledge-based risk factors from an EHR repository.

In PCE, there are 9 risk factors: *gender*, *age*, *race*, total cholesterol (*tc*), high-density lipoproteins cholesterol (*hdl-c*), systolic blood pressure (*sbp*), receiving treatment for high blood pressure (*hbp-treated*), *diabetes* and *smoker*. Referring to our EHR data schema, *gender* and *age* are demographic information, as documented in the Patient table. The LabTest table has values of *tc* and *hdl-c*, while values of *sbp* and *smoker* need to query the FollowUp table. Although there is a PhyExam table, it documents patient image examination results but not including blood pressure values. In the Medication table, drug information has been documented, which provides the value of *hbp-treated*. Since that PCE were developed among Caucasian and African American men and women, the *race* in PCE is "African American" or "White or other". Obviously, in our cohort, we set "White or other" for the *race*. Finally, we set "True" for the factor of *diabetes*, due to our cohort consisting of diabetic patients.

Again, different from registry data, in an EHR data repository, we need to extract the features from sequences of encounters in the observation time window of 1 year before the index event (i.e. T2DM diagnosis firstly observed). Table 1 describes our extraction principles and the instance numbers. For the 54,482 patients as defined above, there is no missing value for *gender* and *age*. However, for the lab test of *tc* and *hdl-c*, only 12.8% and 8.7% have values.

By consultation with local domain experts, these low rates were not surprising, due to the poor patient adherence in China. Interestingly, our EHR data repository is from a tier II city in China, which is a pilot city of hierarchical medical systems. This reflects to our cohort with a high rate of follow up visits in community healthcare centers (CHCs), resulting in 45.6% has "*sbp*" values and 39.9% has "*smoker*" values – it might have null values for columns of systolic_bp and daily_smoking even if there was a follow up visit. Finally, for the treatment for high blood pressure, we map drug names into the high level drug classes, and find those drug names belonging to antihypertensive drug classes. We set "True" for this treatment feature, if an individual was observed having antihypertensive drug classes, else "False" – so, there is no missing value for *hbp_treated*.

**Table 1.** Knowledge-based risk factor extraction principles in EHR and the statistics numbers.

| Risk factor | EHR table | EHR column | Extraction principle | Number |
| --- | --- | --- | --- | --- |
| Gender | Patient | gender | / | 54,482 |
| Age | Patient | birthday | (IndexEvent.commit_time – Patient.birthday)/365 | 54,482 |
| Total Cholesterol | LabTest | test_item_name test_value | For test_item_name = 'tc' and 0 <=(IndexEvent.commit_time – LabTest.commit_time) <= 365 | 6,986 |
| HDL Cholesterol | LabTest | test_item_name test_value | For test_item_name = 'hdl-c' and 0 <= (IndexEvent.commit_time – LabTest.commit_time) <= 365 | 4,722 |
| Systolic BP | FollowUp | systolic_bp | 0 <= (IndexEvent.commit_time – FollowUp.commit_time) <= 365 | 24,827 |
| Treatment for HBP | Medication | drug_name | For drug_name in Antihypertensive drug class and 0 <= (IndexEvent.commit_time – Medication.commit_time) <= 365 | 54,482 |
| Smoker | FollowUp | daily_smoking | 0 <= (IndexEvent.commit_time – FollowUp.commit_time) <= 365 | 21,713 |

Considering lab tests are value sensitive, local domain experts do not recommend us to do data imputation. Instead, they prefer to strictly build the instances. Thus, only 4,143 instances (without any null value) are remained.

**Knowledge-based risk score computation.** The PCE algorithm is publically available, consisting of cox proportional hazards regression models. We implement it to calculate the risk scores for each patient in our cohort. From the official report, PCE has been validated in the population in Caucasian American men and women with AUC of 0.75 and 0.80, while in African American men and women with AUC of 0.71 and 0.82, respectively[14]. The PCE model has also been evaluated in the Korean Heart Study[22], and it reports AUC of 0.72. However, when applying it in our cohort, the AUC is only 0.653. By consultation with domain experts, there could be two reasons for the poor AUC in our cohort. First, PCE is a general ASCVD risk model where diabetes is one of the risk factors, while our cohort consists of diabetic patients. Second, the public PCE algorithm is used to predict 10-year risk, while the index period of our cohort has only 44 months.

**Data-driven feature construction.** The goal of our work is to develop an ASCVD risk prediction model in diabetes based on localized instances. Besides the above knowledge-based risk factors, we would like to leverage more data-driven features to improve the model prediction performance and discovery new risk factors from data insights.

Since that a regional EHR data repository consisting of longitudinal patient visits, the disease history for a given patient is traceable, and possibly, more reliable than the patient's subjective answers to questionnaires. In this respect, we extract the ICD-10 codes from sequences of encounters to construct a feature vector of the disease history. By statistics, there are 27,258 distinct ICD-10 codes (e.g. E11.901 and E11.902) occurring in the Encounter table. Shall we build a vector of 27,258 features? No, because as mentioned above, the usage of ICD-10 is not as regularized as expected, resulting in an extremely sparse feature vector.

One solution is using the 22 chapters of ICD-10 codes to define our data-driven feature vector, where each chapter represents a disease class (e.g. Chapter 4 is endocrine, nutritional and metabolic diseases, with blocks of E00-E90). The other solution is an extraction of the first 3-digit of an ICD-10 code, which represents a disease concept (e.g. E11 is type 2 diabetes mellitus), and we extract 822 distinct 3-digit ICD-10 codes in our cohort. This vector with 822 features will be next processed by feature selection.

We do not involve more clinical measurements (i.e. lab tests and physical examinations) in our study, that's because: (1) at model development phase, it's hard to guarantee the population of interest having done all kinds of lab tests and physical examinations in a retrospective study of EHR; (2) at model application phase – since that our ultimate vision is to take the chronic disease risk prediction model into clinical practice – how to get all of these feature values? If we add more clinical measurements into the model, then it requires these tests or examinations to be done, which most possibly increases the economic burden (even, invasive hurt) of patients.

**Data-driven feature selection.** In machine learning, there are three main supervised feature selection strategies: filter, wrapper and embedded optimization. The methods of wrapper utilize a specific classifier (e.g. logistic regression) to select the subset of features that provides the best performance for a specific metric (e.g. AUC), while the methods of embedded optimization incorporate feature selection directly into the learning process of a model. Towards a fair comparison of learning models (with/without knowledge injection), we do not employ these model-related feature selection methods in this paper, and instead, we use the model-independent filter methods. By calculating scores to represent the relevancy of a feature against the outcome, we select those mostly relevant features and filter others.

**Model learning.** In this study, we emphasize the knowledge injection to model learning process, and as mentioned in Introduction, we will explore three different injection methods. Formally, given an instance $i$ ($1<=i<=n$), the outcome label is denoted as $y_i$ (1 for true, 0 for false), the data-driven feature vector is denoted as $\overline{x_i} = (f_i^1, \ldots, f_i^m)$, where each feature $f^j$ ($1<=j<=m$) represents the value extracted from EHR (e.g. gender, age, the disease history), and the knowledge-based score vector is denoted as $\overline{z_i} = (s_i^1, \ldots, s_i^k)$, where each score $s^j$ ($1<=j<=k$) represents the score calculated from a well-established risk model (e.g. PCE). The formalization of three injection methods is presented below, where $l$ denotes the loss function that is typically cross entropy in classification tasks, $\theta$ represents the learning weights.

- Knowledge injection to input features: Besides data-driven features ($f^1, \ldots, f^m$), we add the knowledge-based scores ($s^1, \ldots, s^k$) as the new features, together inputting to learning function $\sigma$, where the objective function is defined as: $argmin_\theta \frac{1}{n}\sum_{i=1}^n l(y_i, \sigma_\theta(\overline{x_i}, \overline{z_i}))$.

- Knowledge injection to objective functions: We leverage the calculated knowledge-based scores as the supervised signals for model learning, and there are two implantation methods.

    o Teacher-student networks (TSNNs)[9]: Given the $j^{th}$ ($1<=j<=k$) knowledge model, TSNNs are built iteratively, where the teacher network $\rho$ is obtained by projecting the student network $\sigma(\overline{x_i})$ to a knowledge-regularized subspace $s_i^j$, and the student network $\sigma$ is updated to balance between emulating the teacher's output $\rho(\overline{x_i})$ and predicting the outcome labels $y_i$. In particular, after convergence, both the teacher network and the student network could be used for prediction tasks. The objective function of teacher network is $argmin_\theta \frac{1}{n}\sum_{i=1}^n l(\sigma_\theta(\overline{x_i}), \rho_\theta(\overline{x_i})) + \pi_T l(s_i^j, \rho_\theta(\overline{x_i}))$ and the objective function of student network is $argmin_\theta \frac{1}{n}\sum_{i=1}^n l(\rho_\theta(\overline{x_i}), \sigma_\theta(\overline{x_i})) + \pi_S l(y_i, \sigma_\theta(\overline{x_i}))$, where $\pi_T$ and $\pi_S$ are weighting parameters for teacher and student networks respectively.

    o Knowledge-enhanced neural network (KENN)[12]: KENN will be trained end-to-end, in a simultaneous manner, with three network components. Given the $j^{th}$ ($1<=j<=k$) knowledge model, one is a knowledge representation network with target of the calculated score $s_i^j$, the other is a data representation network, and the output layers of these two upper networks will be joined side by side as the input for the final bottom network, with target of the outcome label $y_i$. The objective function is defined as: $argmin_\theta \frac{1}{n}\sum_{i=1}^n (1-\pi)l(y_i, \sigma_\theta(\overline{x_i})) + \pi l(s_i^j, \sigma_\theta(\overline{x_i}))$, where $\pi$ is the weighting parameter to adjust the relative knowledge importance.

- Knowledge injection to output labels: The knowledge-based scores ($s^1, \ldots, s^k$) will be used for decision fusion[11]. Suppose $s^0 = \sigma(\overline{x_i})$ be the score from a data-driven learning model, the fusion function is denoted as $\omega(s_i^0, s_i^1, \ldots, s_i^k)$, and it could be as simple as majority voting, weighted average, or the meta-learning function $\varphi$ with an objective function as defined: $argmin_\theta \frac{1}{n}\sum_{i=1}^n l(y_i, \varphi(\overline{z_i}))$, where $\overline{z_i} = (s_i^0, s_i^1, \ldots, s_i^k)$.

**Model evaluation.** In this study, all models are trained on 80% of the cohort, remaining 20% for testing. The evaluation measure is AUC, considering that we aim at a practicable risk prediction model.

**Results**

For experiments, we start from a raw regional EHR repository including 190,431 diabetic patients, and as described in the section of cohort definition, only 54,482 patients were satisfied by the inclusive criterion. We employ PCE as the knowledge guidance, and due to the badly missing values of clinical measurements, we finally have 4,143 patients for building our ASCVD risk prediction model. Its Kaplan-Meier curve for survival analysis is shown in Figure 5, where 1,535 patients (37.1%) were observed by ASCVD diagnosis, while 2608 patients (62.9%) were censored.

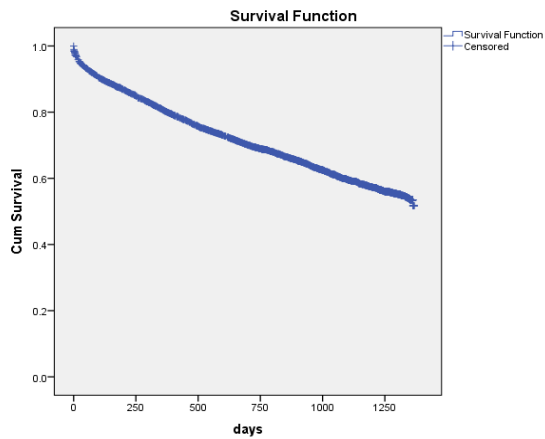

**Figure 5.** Kaplan-Meier curve for survival analysis.

**AUC comparison.** Directly using the PCE algorithm, we calculate the risk scores for our 4,143 patients, receiving AUC of 0.653 – that's our baseline. To develop a more precision ASCVD risk model to well fit the local population, we adopt 8 different algorithms: (1) logistic regression LR, (2) LR extended with knowledge features LR-K, (3) neural network NN, (4) NN extended with knowledge features NN-K, (5) the teacher neural network TSNN-T, (6) the student neural network TSNN-S, (7) the knowledge-enhanced neural network KENN, and (8) the decision fusion DF using weighted average.

For implementation, LR and LR-K are developed using the sklearn library[23], while neural networks (for NN, NN-K, TSNN and KENN) are all developed using the keras library[24]. In particular, TSNN is developed with two iteratively trained neural networks, and we set the hyper-parameter of $\pi_T$ as 0.653 (which reflects our PCE knowledge performance of AUC=0.653) while $\pi_S$ as 1. KENN is developed with three simultaneously trained neural networks, and similarly, we set the hyper-parameter of $\pi$ as 0.653. For each neural network building, we use 3 dense layers (i.e. fully connected layers), and to avoid overfitting, we stack a dropout (of 0.5) layer after each dense layer. These neural networks are all trained for 50 epochs, and considering the relatively small size of our instances, all hidden layers are set up with 8 cells and the batch size is 8.

Table 2 shows the model comparison. All models are trained on 80% of the cohort (i.e. 3301 instances), remaining 20% for testing (i.e. 842 instances). We have four experiments: (EX-1) using 7 known risk factors as defined in PCE, (EX-2) adding 22 chapters of ICD-10 codes, (EX-3) adding 822 distinct 3-digit ICD-10 codes, and (EX-4) adding 20 highest scoring 3-digit ICD-10 codes as automatically selected by applying a univariate filter (the p-value from chi-squared test). From the experimental results, we have the following findings.

1. AUC is improved with more features involved, and for instance, LR has AUC of 0.690 for testing in EX-1, while LR has AUC of 0.715 for testing in EX-3, improved by 3.6%

- However, facing the raw high dimension features (e.g. 822 features in EX-3), the performance is unacceptable (even worse than the baseline AUC 0.653), and after feature selection (i.e. EX-4), the performance becomes better

2. Due to the limited training data size, NN is not promising, compared with LR, and worse, NN is more prone to be overfitting than LR – for every experiment, in training, NN has a higher AUC than LR

3. The knowledge (PCE scoring) guidance is not strong, resulting a slight improvement of knowledge injection methods – for instance, neither LR-K significantly outperforms LR, nor NN-K outperforms NN

4. Performance of TSNN-S and KENN are roughly the same, achieving the highest AUC in every experiment, and the final best model is using either TSNN-S or KENN with 7 known risk factors plus 20 selected data-driven features, with AUC of 0.723 – compared with the PCE baseline of 0.653, improved by 10.7%

Table 2. AUC comparison of LR, LR-K, NN, NN-K, TSNN-T, TSNN-S, KENN and DF-WA.

|  | EX-1: 7 known risk factors | | EX-2: 7 known risk factors + 22 ICD chapters | | EX-3: 7 known risk factors + 822 ICD 3-digit codes | | EX-4: 7 known risk factors + 20 ICD 3-digit codes after feature selection | |
|---|---|---|---|---|---|---|---|---|
|  | Training | Testing | Training | Testing | Training | Testing | Training | Testing |
| LR | 0.719 | 0.690 | 0.736 | 0.715 | 0.873 | 0.638 | 0.755 | 0.722 |
| LR-K | 0.719 | 0.690 | 0.736 | 0.715 | 0.874 | 0.638 | 0.756 | 0.722 |
| NN | 0.729 | 0.691 | 0.768 | 0.717 | 0.885 | 0.674 | 0.785 | 0.720 |
| NN-K | 0.736 | 0.691 | 0.771 | 0.718 | 0.887 | 0.678 | 0.789 | 0.721 |
| TSNN-T | 0.648 | 0.644 | 0.649 | 0.648 | 0.619 | 0.636 | 0.643 | 0.643 |
| TSNN-S | 0.722 | **0.692** | 0.738 | 0.718 | 0.772 | **0.703** | 0.760 | **0.723** |
| KENN | 0.714 | **0.692** | 0.737 | **0.719** | 0.774 | 0.702 | 0.751 | **0.723** |
| DF-WA | 0.715 | 0.683 | 0.733 | 0.708 | 0.871 | 0.637 | 0.752 | 0.716 |

Table 3. Univariate analysis and cox regression analysis.

| 7 known risk factors + 22 ICD chapters | | | | | | | 7 known risk factors + 20 ICD codes after feature selection | | | | | | |
|---|---|---|---|---|---|---|---|---|---|---|---|---|---|
|  | Univariate | | Cox regression analysis | | | |  | Univariate | | Cox regression analysis | | | |
|  | Sig. | Pearson | Sig. | Exp(B) | 95.0% CI for Exp(B) | |  | Sig. | Pearson | Sig. | Exp(B) | 95.0% CI for Exp(B) | |
|  |  |  |  |  | Lower | Upper |  |  |  |  |  | Lower | Upper |
| **gender** | **.000** | **.099** | **.000** | **1.235** | 1.109 | 1.375 | **gender** | **.000** | **.099** | **.000** | **1.214** | 1.091 | 1.352 |
| **age** | **.000** | **.313** | **.000** | **1.035** | 1.030 | 1.039 | **age** | **.000** | **.313** | **.000** | **1.032** | 1.027 | 1.036 |
| tc | .712 | .006 | .570 | .987 | .942 | 1.033 | tc | .712 | .006 | .736 | .992 | .947 | 1.039 |
| hdl-c | **.001** | **.052** | .435 | 1.064 | .911 | 1.243 | hdl-c | **.001** | **.052** | .988 | 1.001 | .858 | 1.169 |
| **sbp** | **.000** | **.135** | **.000** | **1.005** | 1.003 | 1.008 | **sbp** | **.000** | **.135** | **.001** | **1.005** | 1.002 | 1.008 |
| **treated** | **.000** | **.205** | **.000** | **.716** | .639 | .802 | **treated** | **.000** | **.205** | **.044** | **.880** | .776 | .997 |
| smoker | .211 | -.019 | .881 | .976 | .708 | 1.345 | smoker | .211 | -.019 | .669 | .932 | .677 | 1.285 |
| c1 | **.021** | **.036** | .472 | .987 | .954 | 1.022 | **I10** | **.000** | **.225** | **.000** | **1.344** | 1.189 | 1.519 |
| c2 | **.003** | **.045** | .404 | 1.006 | .992 | 1.020 | **E78** | **.000** | **.175** | **.000** | **1.344** | 1.195 | 1.512 |
| c3 | **.028** | **.034** | .613 | .993 | .969 | 1.019 | **H81** | **.000** | **.140** | **.001** | **1.306** | 1.113 | 1.532 |
| c4 | **.000** | **.119** | .983 | 1.000 | .995 | 1.005 | M54 | **.000** | **.110** | .569 | 1.048 | .892 | 1.231 |
| **c5** | **.015** | **.038** | **.020** | **1.017** | 1.003 | 1.031 | H10 | **.000** | **.074** | .368 | 1.105 | .889 | 1.374 |
| c6 | **.000** | **.080** | .698 | 1.007 | .974 | 1.040 | J39 | **.000** | **.099** | .955 | 1.004 | .859 | 1.174 |
| c7 | **.002** | **.049** | .821 | .994 | .943 | 1.048 | M81 | **.000** | **.097** | .956 | .995 | .820 | 1.206 |
| c8 | **.000** | **.080** | .867 | 1.003 | .967 | 1.041 | K29 | **.000** | **.092** | .149 | 1.104 | .965 | 1.264 |
| c9 | **.000** | **.154** | .129 | 1.006 | .998 | 1.015 | M13 | **.000** | **.072** | .274 | 1.130 | .907 | 1.408 |
| **c10** | **.000** | **.153** | **.000** | **1.040** | 1.025 | 1.056 | **G47** | **.000** | **.088** | **.027** | **1.253** | 1.026 | 1.531 |
| c11 | **.000** | **.071** | .803 | .998 | .986 | 1.011 | M25 | **.000** | **.109** | .082 | 1.162 | .981 | 1.376 |
| c12 | .248 | .018 | .228 | .983 | .956 | 1.011 | J40 | **.000** | **.088** | .153 | 1.146 | .951 | 1.381 |
| c13 | **.000** | **.089** | .291 | 1.005 | .996 | 1.013 | **J06** | **.000** | **.085** | **.039** | **1.155** | 1.008 | 1.324 |
| c14 | **.024** | **.035** | .092 | 1.006 | .999 | 1.013 | J00 | **.000** | **.108** | .929 | 1.008 | .846 | 1.201 |
| c15 | **.016** | **-.037** | .595 | .952 | .794 | 1.141 | M24 | **.000** | **.076** | .171 | 1.211 | .921 | 1.592 |
| c16 | .278 | -.017 | .934 | .033 | .000 | .000 | G45 | **.000** | **.086** | .089 | 1.268 | .964 | 1.667 |
| c17 | .404 | -.013 | .509 | .642 | .172 | 2.392 | J31 | **.000** | **.083** | .389 | 1.085 | .901 | 1.307 |
| **c18** | **.000** | **.117** | **.049** | **1.018** | 1.000 | 1.035 | R53 | **.000** | **.076** | .219 | .870 | .696 | 1.087 |
| c19 | .611 | .008 | .114 | .931 | .852 | 1.017 | **R09** | **.000** | **.125** | **.042** | **1.261** | 1.009 | 1.577 |
| c20 | .061 | .029 | **.038** | **1.346** | 1.017 | 1.780 | **J45** | **.000** | **.070** | **.017** | **1.498** | 1.076 | 2.084 |
| c21 | .815 | .004 | .127 | .990 | .977 | 1.003 |  |  |  |  |  |  |  |

**Clinical findings.** Besides the prediction model development, we provide data insights to discover more risk factors for ASCVD in diabetes. Table 3 presents the univariate analysis and the cox regression analysis for models using 7 known risk factors plus 22 ICD chapters (ref. EX-2), and using 7 known risk factors plus 20 ICD 3-digit codes after feature selection (ref. EX-4), respectively. For those features with statistical significance (p-value <= 0.05) in both analysis results, we highlight them in bold.

For the univariate analysis, we use the Pearson correlation. For the 7 known risk factors, either in EX-2 or in EX-4, the Pearson correlation coefficients are the same. Except *tc* and *smoker*, other 5 risk factors have the statistical significance. Particularly, in EX-4, the 20 features of ICD 3-digit codes are selected because of their high relevancy against the outcome, and undoubtedly, they all have the statistical significance. However, for the 22 ICD chapters in EX-2, only c1-c11, c13-c15 and c18 have the statistical significance, where c15 (Pregnancy, childbirth and the puerperium) is the only protective factor. For c22 (Codes for special purposes), we have no observation for this ICD chapter in our cohort, so c22 has been ignored in Table 2.

For the cox regression analysis, in both EX-2 and EX-4, *hdl-c*, *tc* and *smoker* are not statistically significant, while *sbp-treated* is the only protective factor with statistical significance. For the 22 ICD chapters in EX-2, only c5 (Mental and behavioral disorders), c10 (Diseases of the respiratory system), c18 (Symptoms, signs and abnormal clinical and laboratory findings, not elsewhere classified) and c20 (External causes of morbidity and mortality) have the statistical significance, while for the selected 20 ICD codes in EX-4, only I10 (Essential (primary) hypertension), E78 (Disorders of lipoprotein metabolism and other lipidemias), H81 (Disorders of vestibular function), G47 (Sleep disorders), J06 (Acute upper respiratory infections of multiple and unspecified sites), R09 (Other symptoms and signs involving the circulatory and respiratory system) and J45 (Asthma) have the statistical significance.

**Discussion**

To echo the question in Introduction, we have confidence to answer "YES", esp., after our exploration on developing a knowledge-enhanced ASCVD risk model in diabetes from a raw EHR repository. However, from this case study, we also realize it deserves more investigations towards a more practicable risk model for clinical decision support.

First, we admit our learning data size is poorly small in this study. Facing the original 190,431 diabetic patients, why do we finally only get 4,143 instances? We have to own this to the data quality of our EHR repository. On the one hand, the standardized usage of medical code systems (e.g. ICD-10) is not satisfactory in developing countries (e.g. China), and to address this problem, we are planning to learn high-level medical concepts from free text[20] in our future work. On the other hand, the regular collection of clinical measurements (e.g. lab tests) is seriously problematic, mostly due to the low quality of patient adherence. Although regularly evaluating the fasting lipid profile has been recommended in diabetic guidelines[18], only 10% diabetic patients have fasting lipid profiles in our EHR repository. Similar ratio even for HbA1c, despite it recommends to test HbA1c within 3 months for diabetic patients[18]. Besides, by consulting local clinicians, we were told that portable tests/examinations such as blood glucose and blood pressure, most possibly, totally not documented in the hospital information system (neither existing in the laboratory information system). That's why we only use the disease history as data-driven features in this study. Actually, using machine learning methods for EHR data analysis, related work mostly takes the disease history and medication history into account[3,4], disregarding the clinical measurements. The exception is using EHR data in the ICU (intensive care unit)[5], which clinical measurements are always continuously observable. In our study, to avoid the strong correlation between the disease history and the medication history, we finally extract the disease history as data-driven features.

Second, the knowledge guidance signal is a bit weak. Our motivation is using knowledge guidance to greatly improve the prediction performance, but our experimental results show a slight improvement. Besides, more advanced methods of knowledge injection to objective functions are desirable. Our implementation of TSNN/KENN requires the instance-level knowledge labels (i.e. the calculated risk score for each instance), while the structure-level knowledge injection[10] is more attractive. In clinical scenario, it might be directly projecting the logical rules (e.g. clinical guidelines for treatment recommendation) or the arithmetic formulas (e.g. logistic/cox regression for risk prediction) into the neural network learning.

**Conclusion**

With the popularity of building regional healthcare information platform[2,15], EHR repositories have stored (billions of) clinical documents for (millions of) patients in the past decade (in terms of a tier II city in China). Towards the meaningful use of EHR, this paper proposes to develop the knowledge-enhanced chronic disease risk prediction models from regional EHR repositories. The knowledge enhancement is meant to leverage those well-established

(inter-)national risk models for guidance in the EHR data learning process, and chronic diseases fit well with longitudinal patient visits documented in regional EHR repositories.

As a case study, we present our experiments on developing an ASCVD risk model in diabetes, from the cohort definition and feature construction, to the knowledge injection and model evaluation. Specially, we implement different algorithms for knowledge injection into input features, objective functions and output labels. The experimental results demonstrate the prediction performance improvement of our models, and by consultation with domain experts, our models also provide data insights for clinical findings. This paper exploits the way for knowledge learning symbiosis in healthcare, and for future work, we are planning to validate our methods on more EHR repositories to address different clinical problems.